\documentclass[letterpaper, 10 pt, conference]{ieeeconf}  

\IEEEoverridecommandlockouts                              

\usepackage{amsmath}
\usepackage{amsmath,amssymb}
\usepackage{booktabs}
\usepackage{graphicx} 
\usepackage{algorithm}
\usepackage{algpseudocode}
\usepackage{xcolor}
\usepackage{wrapfig}
\usepackage{multirow}
\usepackage{tabularx}
\usepackage{adjustbox}
\usepackage{cite}
\usepackage[dvipsnames]{xcolor}
\usepackage{url}

\newcommand{\gaitok}[1]{\textcolor{ForestGreen}{#1}}
\newcommand{\gaitbad}[1]{\textcolor{BrickRed}{#1}}

\title{\LARGE \bf
Learning Gait-Aware Quadruped Locomotion with Temporal Logic Specifications
}

\author{%
 Merve Atasever \quad Keyan Azbijari \quad Cagan Bakirci \\ Alfredo Reina Corona  \quad
Tolga Izdas \quad Jyotirmoy V.~Deshmukh \\
 University of Southern California, Los Angeles, California, USA \\
 \texttt{\{atasever,azbijari,cbakirci,reinacor,izdas,jdeshmuk\}@usc.edu}
 }

\begin{document}

\maketitle
\thispagestyle{empty}
\pagestyle{empty}

\begin{abstract}
Reinforcement learning (RL) for quadruped locomotion commonly depends on fixed,
hand-crafted, and Markovian reward functions that may limit interpretability of
learned policies and may lack explicit control over gait behaviors. We introduce a
framework where distinct gaits are specified using parameterized constraints
expressed in Signal Temporal Logic (STL). These include safety bounds, gait
synchronization constraints, command tracking, and actuation bounds. From these
specifications, we develop a reward shaping mechanism that provides learning
agents a dense, continuous reward landscape that encodes desired behavior. We
define parametric STL templates for three speed regimes (walking-trot, trot,
bound), calibrate their parameters from reference rollouts, and compute rewards
from using smooth approximations of STL robustness over the rollouts. The
generated rewards can be used to provide shaped gradients compatible with
Proximal Policy Optimization (PPO). We instantiate the approach on Google's
Barkour quadruped robot in MuJoCo XLA (MJX). We use parallelization within the
simulator to improve training speeds and use domain randomization to robustify learned policies. Compared with hand-crafted rewards, an expert-switching oracle, and Text2Reward, Human-STL maintains high command-tracking success across the evaluated
speed range while exhibiting substantially higher consistency with the intended speed-dependent gait structures. Videos can be found on our project website: \\
\url{https://stl-locomotion.github.io/}.

\end{abstract}

\section{Introduction}


Legged robots, including both bipeds and quadrupeds, have made remarkable progress in recent years, with modern platforms exhibiting agile walking, crawling, running, and more in increasingly challenging settings \cite{minicheetah, learningquadimit, learningquadsci}. Despite this progress, achieving animal-level robustness across diverse speed regimes and gait patterns
remains an open challenge. Classical model-based control approaches, such as differential dynamic programming (DDP) and model predictive control (MPC), have enabled impressive quadrupedal behaviors by explicitly optimizing trajectories under dynamics and contact constraints. However, their performance often depends on accurate system models and carefully designed cost functions, all of which can be difficult to obtain on real hardware. \cite{mpc, di2018dynamicmpc}. Reinforcement learning (RL) offers a complementary path by directly optimizing closed-loop policies in simulation, and recent advances in high-fidelity physics engines and standardized legged-locomotion platforms, such as MuJoCo, Isaac Gym, Google Barkour, Unitree A1/Go1, and ANYmal, have further accelerated progress in this direction.\cite{todorov2012mujoco, anymal, makoviychuk2021isaac, caluwaerts2023barkour}. 

A key requirement for quadrupedal locomotion is the ability to operate across multiple speed regimes by switching between distinct gait patterns. For example, walking, trotting, and bounding differ not only in velocity but also in their contact timing, duty factors, and employed support structures. Locomotion controllers must therefore satisfy several objectives simultaneously: accurate command tracking over a wide range of speeds, safety and stability under disturbances, and structured gait behavior whose contact patterns remain appropriate for the commanded regime. State-of-the-art deep RL pipelines address the first objective (and partially the second) through carefully engineered reward terms, curriculum learning, and domain randomization \cite{tan2018sim, anymal, caluwaerts2023barkour}. However, these reward functions are often difficult to interpret, and they provide only indirect control over contact-sequence structure and speed-dependent gait transitions. This limitation becomes especially important in multi-gait locomotion. Heuristic rewards based on forward velocity, energy consumption, and posture regularization, can be sufficient for learning a single dominant gait such as trotting; yet, the same static rewards may fail to induce distinct gait regimes or smooth transitions. 
Existing approaches often address this issue through imitation learning or reference-based objectives, using motion-capture data, or hand-designed gait schedules to impose desired locomotion patterns \cite{inverserl,li2023fastmimic}.

\begin{figure}[t]
  \centering
  \includegraphics[width=0.4\textwidth]{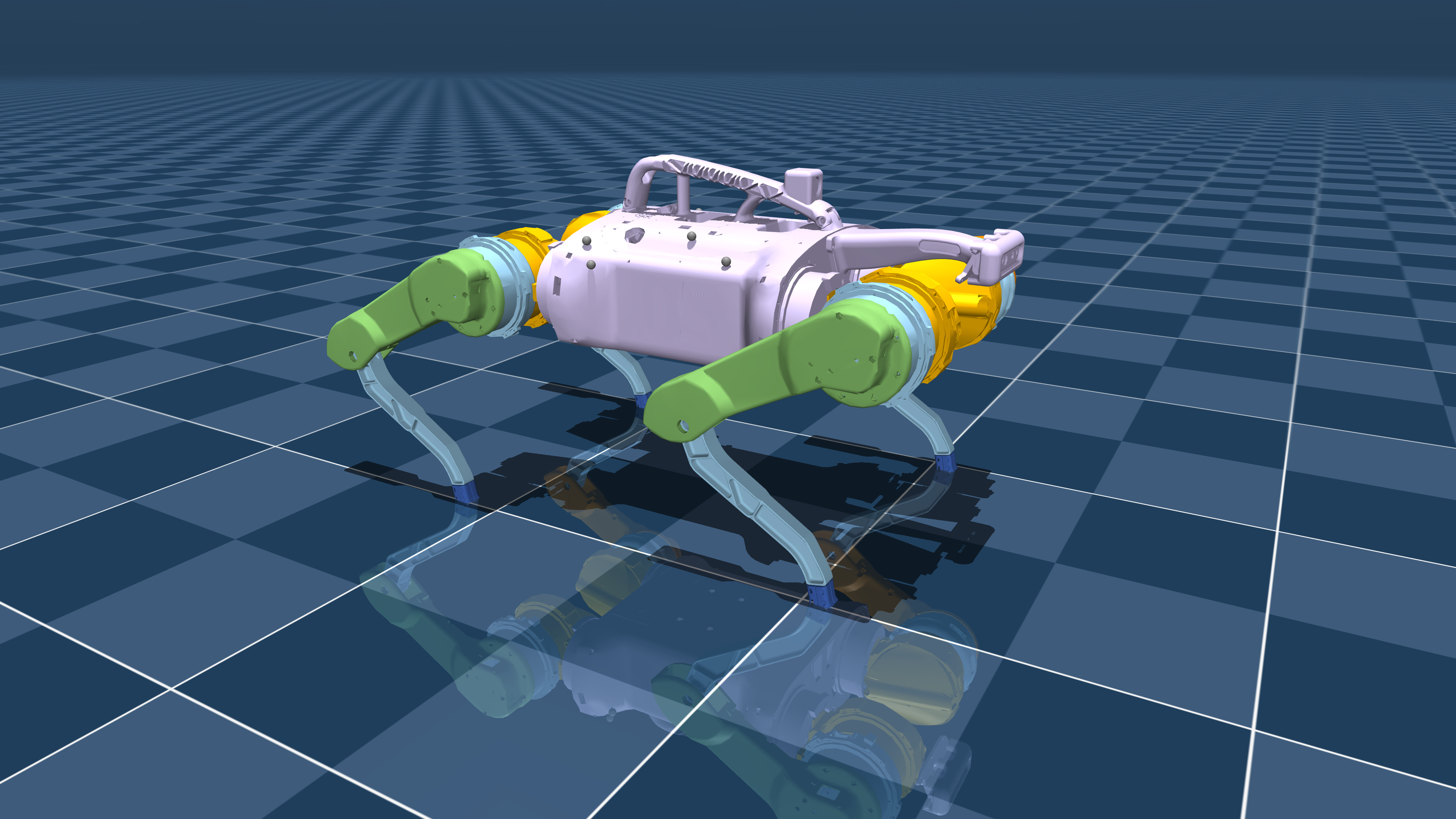}
  \vspace{-5pt}  
  \caption{Barkour vb robot in MJX.}
  \label{fig:barkour}
  \vspace{-15pt}                       
\end{figure}

\begin{figure*}[t]
    \centering
    \includegraphics[width=0.75\linewidth]{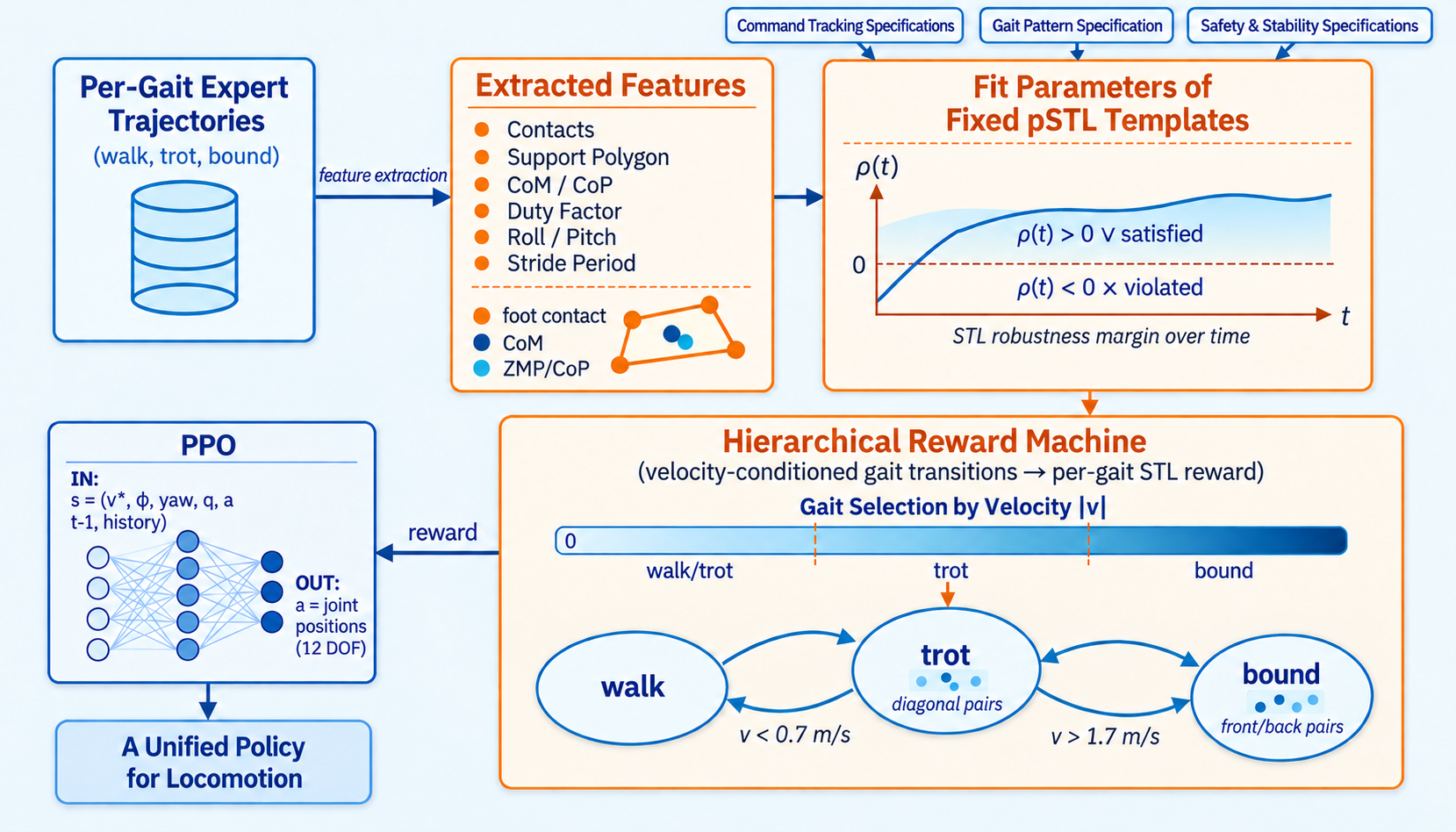}
    \caption{Overall Pipeline.}
    \label{fig:pipeline}
     \vspace{-15pt} 
\end{figure*}

Formal specifications provide an alternative perspective: rather than expressing locomotion objectives through loosely coupled rewards, desired behaviors can be encoded as logical specifications. Signal Temporal Logic (STL) is a framework that has been used in robotics to express bounded-time constraints over real-valued signals. It admits quantitative or {\em robustness} semantics, i.e. signed margins indicating how well a trajectory satisfies a specification \cite{maler2004monitoring}. 
 Prior work has demonstrated that robustness can be used as a reward to accelerate RL and improve interpretability \cite{ding2014optimal, li2017reinforcement, aksaray2016q, deshmukh2017robust, balakrishnan2019structured, jiang2021temporal, hasanbeig2020deep, wen2015correct, icarte2022reward}. We build on this using STL robustness to encode multi-gait coordination and safety requirements. We target a key engineering challenge in quadrupedal locomotion: learning a single policy that adapts its gait structure across speed regimes (walking-trot $\to$ trot $\to$ bound). We specifically use mode-conditioned STL templates (where template parameters are mined from expert trajectories), and the shaped reward chooses the appropriate template based on the commanded speed regime. This design is motivated by: (1) many locomotion requirements,  including footfall pattern and stability, are inherently temporal: they describe behavior over short trajectory segments rather than single states, and (2) switching between gait-specific objectives near speed thresholds can introduce reward discontinuities and destabilize learning unless the mode logic is designed carefully (e.g., hysteresis around regime boundaries). Each reward component corresponds to a human-readable requirement, active gait template changes with the commanded speed regime. 
As robustness values for the specifications provide diagnostic signals after training, the same STL structure enables failure attribution: poor rollouts can be analyzed in terms of violated predicates, such as insufficient support, actuator-limit violations, or poor orientation. This specification-level feedback is difficult to obtain from monolithic rewards and provides a practical tool for debugging multi-gait locomotion policies.

\noindent{\bf Main Contributions.}
(1) A continuous STL-robustness reward formulation for gait-aware quadruped locomotion, where speed-conditioned gait structure, tracking, and safety are expressed over finite temporal windows rather than only through step-wise reward terms.
(2) Data-driven parameter learning pipeline that fits a set of parametric gait templates from expert trajectories, producing interpretable specs aligned with gait statistics.
(3) A unified single policy that supports multiple speed-conditioned gait regimes while maintaining high command-tracking success across the evaluated velocity range.
(4) An LLM-generated STL ablation that evaluates temporal logic as an interpretable intermediate representation between language-based reward design and policy learning \cite{xie2024text2reward}.

\noindent{\bf Related Work.}

\noindent {\em Reward Shaping:} As RL becomes standard in locomotion, reward design remains a key bottleneck \cite{ eschmann2021reward, icarte2022reward}. Most work relies on hand-crafted weighted rewards with empirically tuned coefficients, justified through ablations rather than first principles \cite{ eschmann2021reward}. Alternatives such as inverse RL and preference-based RL infer rewards from demonstrations or human feedback, but require expert trajectories or annotations and introduce additional optimization overhead \cite{inverserl, rlhf, youm2023imitating, li2023fastmimic}. (Our approach also uses expert trajectories, but only to calibrate a set of symbolic thresholds.) Recent LLM-based reward-design methods generate executable dense reward functions directly from natural-language task descriptions \cite{xie2024text2reward,ma2024eureka}, but struggle with high-speed locomotion. Finally, prior work has explored logic-guided RL by converting logic objectives into quantitative rewards (e.g., LTL/STL reward shaping) and learning temporal logic specifications from demonstrations, often in navigation/manipulation settings  \cite{hasanbeig2020deep, li2017reinforcement, stlrl, stlrl3}. Formal specifications have also guided planning and control for legged robots, including locomotion over cluttered terrain and STL-informed MPC for bipedal push recovery \cite{gu2025robust}. However, comparable formulations for quadrupeds remain limited. Existing logic-driven gait learning for quadrupeds typically uses discrete logic rules or reward machines over foot-contact propositions, enabling structured gait sequencing but operating mainly on contact-level logic and step-wise signals \cite{gaitstrategiesnature, defazio2024learning}.

\noindent {\em Quadruped Locomotion:} Work in biology inspired locomotion focuses on gait expression and transitions emerging from structured interlimb coordination and speed-dependent dynamics rather than control outcomes \cite{schoner1990synergetic, danner2016central}. This has shaped research using central pattern generator (CPG) models and sensory-feedback frameworks used to encode rhythmic structure and transition behavior \cite{humphreys2023bio, neunert2017trajectory, bellegarda2025allgaits}. While effective, this work typically depends on accurate dynamic models, incurs computational cost, and needs carefully engineered optimization pipelines.
More recent work focuses on \emph{hierarchical RL, phase-guided control, and unified multi-gait representations} to achieve multi-gait locomotion and gait transitions \cite{hu2019learning, tsounis2020deepgait, shao2021learning,xu2022learning}. One hierarchical framework uses a high-level gait policy to specify footfall patterns with low-level MPC-based motion tracking, yielding energy-efficient gait transitions as speed increases \cite{yang2022fast}. Related work argues that single end-to-end locomotion policies often collapse to one dominant gait and instead proposes hierarchical controllers capable of generating pace, trot, and bound while tracking velocity commands \cite{kim2021learning}. Other approaches seek unified latent representations rather than explicit switching or hand-designed phase templates. These learn  compact embeddings that capture shared structure across gait families and interpolate between them \cite{mitchell2024gaitor, shafiee2024manyquadrupeds}. These methods reduce the need for manually enumerated gait templates, but their learned representations can be less interpretable than structured controllers with explicit gait objectives.

Overall, existing multi-gait methods primarily structure locomotion through hierarchical controllers, phase-based representations, or learned gait embeddings, while existing logic-driven quadruped approaches operate mainly on discrete contact propositions and step-wise logical signals. Our formulation instead uses continuous STL robustness over real-valued locomotion and contact signals as the reward interface itself. This enables temporally extended, mode-conditioned objectives whose numerical parameters are grounded from expert trajectories and whose predicate-level robustness remains available for post-hoc diagnosis. Thus, our contribution is not multi-gait control alone, but a specification-driven reward representation that jointly expresses temporal gait structure, command tracking, and safety within a unified policy.

\section{Technical Background and Preliminaries}

\noindent{\bf Problem Setting.}
We consider Google's Barkour vb in MJX \cite{caluwaerts2023barkour,todorov2012mujoco}. At each control step ($\Delta t=0.02$ s), the policy receives commanded linear and angular velocities $(v_x^{cmd},v_y^{cmd},\omega_z^{cmd})$, previous action, an observation history of 15 time steps for robot proprioception, and a mode flag (walk/trot/bound), and outputs 12 normalized joint-position commands. A PD controller maps these commands to actuator torques. The goal is to learn a single policy $
\pi_\theta(a_t \mid o_t),
$ where $\theta, a_t, o_t$ denotes the current policy parameters, action, and observation respectively, that achieves accurate command tracking while exhibiting stable, interpretable, and mode-appropriate gait behavior across multiple speed regimes. 

\noindent{\bf Mode-Dependent Gait Selection.}
Using the same gait for low speed and high speed motion may be inefficient or infeasible for quadrupeds \cite{fu2021minimizing, yang2022fast}. To determine transition velocity thresholds between speed regimes, we leverage the Froude number which is used to characterize gait transition speeds of different-sized quadrupeds:
\(
\mathrm{Fr} = \frac{v^2}{g h},
\)
where Fr is the Froude number, $v$ is the linear velocity of the robot's torso, $g$ is the acceleration due to gravity, and $h$ is the maximum leg length \cite{alexander1983dynamic}. 
Building on bio-inspired robotics gait design methodologies and reported gait structures that quadrupeds achieved to succeed, we use three speed regimes (modes): walking-trot ($W$), trot ($T$), and bound ($B$), selected based on commanded forward velocity $|v_x^{cmd}(t)|$. 
Using the Barkour robot's leg length of $0.41~\mathrm{m}$ and the reference Froude numbers of $0.15$ and $0.74$ \cite{humphreys2023bio}, we establish the respective gait transition velocity thresholds at $0.7~\mathrm{m/s}$ and $1.7~\mathrm{m/s}$. Here, diagonal legs move together for mid-speed regime (trot), whereas forelegs make contact approximately in phase, followed by the hind legs to achieve high speeds (bound). We denote the active locomotion mode at time $t$ as $g(t)\in\{W,T,B\}$, which also serves as the Boolean indicator for the corresponding gait mode.

\noindent{\bf Signal Temporal Logic (STL) \& Quantitative Robustness.}
Signal Temporal Logic (STL) specifies temporal properties over real-valued signals $x(t)$ using Boolean and temporal operators \cite{maler2004monitoring}.
Formulas are built from atomic predicates $f(x(t)) \ge 0$, Boolean connectives ($\wedge,\vee,\neg$), and time-bounded temporal operators $\mathbf{G}_{[a,b]}$ (always) and $\mathbf{F}_{[a,b]}$ (eventually).
We use quantitative semantics of STL \cite{donze2010robust,fainekos2009robustness}; for a given trace $x$, and each formula $\varphi$ $\rho^\varphi(t)$ assigns a real value capturing the degree of satisfaction.
Positive values imply satisfaction; negative imply violation, and the magnitude measures the satisfaction margin. 
For atomic predicates, $\rho(f(x)\ge0,x,t)=f(x(t))$.
For conjunction/disjunction, robustness uses min/max:
\(
\rho^{\varphi\wedge\psi}(t) = \min(\rho^\varphi(t),\rho^\psi(t))\), and
\(\rho^{\varphi\vee\psi}(t) = \max(\rho^\varphi(t),\rho^\psi(t)).
\)
For time-bounded operators, we have 
\(\rho^{\mathbf{G}_I\varphi}(t)=\inf_{t'\in t+I}\rho^\varphi(t')\),
and 
\(\rho^{\mathbf{F}_I\varphi}(t)=\sup_{t'\in t+I}\rho^\varphi(t')\) for an arbitrary time interval $I=[a,b]$.
Parametric Signal Temporal Logic (PSTL) extends STL by allowing constants in predicates and intervals in temporal operators to be represented by parameters \cite{asarin2011parametric}. The {\em specification mining} problem seeks to infer parameter ranges where the template formula is satisfied providing a data-driven way to instantiate temporal specifications from expert-provided locomotion trajectories. 

\section{Methodology}

\noindent{\bf STL Specifications for Multi-Gait Locomotion.}
We consider three locomotion regimes parameterized by the commanded forward velocity $v_x^{cmd}$:
\emph{walking-trot} for $0 \le v_x^{cmd} \le 0.7$ m/s,
\emph{trot} for $0.7 < v_x^{cmd} \le 1.7$ m/s, and
\emph{bound} for $v_x^{cmd} > 1.7$ m/s. For each mode $g(t)\in\{W,T,B\}$, we define a mode-specific STL specification composed of a shared safety, velocity tracking, and a gait-shape part. During policy training, we use a trailing-window over the realized trajectory prefix. At step \(t\), the quantitative robustness
of the active specification is evaluated over \(W=[\max(0,t-H),t]\), where the window size \(H\) is set to $30$ timesteps to cover roughly one complete gait cycle for slow walking.

For any Boolean event \(A(k)\), we define the window fraction
\(
p_A(t)=\frac{1}{|\mathcal{W}|}
\sum_{k\in\mathcal{W}}\mathbf{1}\{A(k)\}.
\)
Positive robustness indicates satisfaction, negative robustness indicates violation, and the magnitude gives the satisfaction/violation margin. Tables~\ref{tab:shared-stl} and~\ref{tab:gait-stl} summarize the predicates used in our reward.

\begin{table}[t]
\centering
\small
\setlength{\tabcolsep}{2.8pt}
\renewcommand{\arraystretch}{1.2}
\caption{Shared safety and tracking STL predicates.}
\label{tab:shared-stl}
\begin{tabularx}{0.95\linewidth}{@{}c | l X @{}}
\toprule
ID & Predicate & STL constraint \\
\midrule
S1 & Torque limit &
\(\mathbf{G}_{\mathcal{W}}(|\tau_j|\le \tau^{\max}_j),\ \forall j\) \\

S2 & Roll limit &
\(\mathbf{G}_{\mathcal{W}}(|\phi|\le \phi_{\max}^g)\) \\

S3 & Pitch limit &
\(\mathbf{G}_{\mathcal{W}}(|\theta|\le \theta_{\max}^g)\) \\

S4 & CoM height &
\(\mathbf{G}_{\mathcal{W}}(z_{\mathrm{com}}\ge z_{\min}^g)\) \\

S5 & Support contacts &
\(\mathbf{G}_{\mathcal{W}}(n_c\ge n_{\min}^g)\) \\

T1 & Linear velocity &
\(\mathbf{G}_{\mathcal{W}}(|v_i-v_i^{cmd}|\le \epsilon_{v,i}^g),\ i\in\{x,y\}\) \\

T2 & Yaw-rate tracking &
\(\mathbf{G}_{\mathcal{W}}(|\omega_z-\omega_z^{cmd}|\le \epsilon_\omega^g)\) \\
\bottomrule
\end{tabularx}
\vspace{-8pt}
\end{table}



\begin{table}[t]
\centering
\small
\setlength{\tabcolsep}{1.9pt}
\renewcommand{\arraystretch}{1.2}
\caption{Mode-dependent gait-structure STL predicates.}
\label{tab:gait-stl}
\begin{tabularx}{0.98\linewidth}{@{}c | l | l l @{}}
\toprule
ID & Mode & Predicate & STL specifications \\
\midrule

G1 & \(W,T\) & Diagonal phase error &
\(\mathbf{G}_{\mathcal{W}}
(e_{\mathrm{diag}}\le \epsilon_{\mathrm{diag}}^g)\) \\

G2 & \(W,T\) & Diagonal pair fraction &
\(\mathbf{G}_{\mathcal{W}}
(p_{\mathrm{diag2}}\ge \eta_{\mathrm{diag2}}^g)\) \\

G3 & \(W\) & Multi-support moment &
\(\mathbf{G}_{\mathcal{W}}
\mathbf{F}_{[0,K]}(n_c\ge 3)\) \\

G4 & \(T\) & Two-leg support &
\(\mathbf{G}_{\mathcal{W}}
(p_{2\mathrm{contact}}\ge p_{2,\min})\) 
\\

B1 & \(B\) & Aerial-phase occupancy &
\(\mathbf{G}_{\mathcal{W}}
(p_{\mathrm{flight}}\ge p_{\mathrm{flight},\min})\) \\

B2 & \(B\) & Pair-only support &
\(\mathbf{G}_{\mathcal{W}}
(p_{\mathrm{front}}\ge p_{\mathrm{front},\min})\) \\ 

B3 & \(B\) & Pair-only support &
\(\mathbf{G}_{\mathcal{W}}
(p_{\mathrm{hind}}\ge p_{\mathrm{hind},\min)}\) \\

B4 & \(B\) & Four contact fraction &
\(\mathbf{G}_{\mathcal{W}}
(p_4\in[p_{4,\min},p_{4,\max}])\) \\

B5 & \(B\) & Trot suppression &
\(\mathbf{G}_{\mathcal{W}}
(p_{\mathrm{diag2}}\le p_{\mathrm{diag2},\max})\) \\

B6 & \(B\) & Pair touchdown events &
\(\mathbf{F}_{\mathcal{W}}E_F
\wedge
\mathbf{F}_{\mathcal{W}}E_H\) \\

\bottomrule
\end{tabularx}
\vspace{-15pt}
\end{table}

Using the predicate definitions in Tables~\ref{tab:shared-stl}-\ref{tab:gait-stl}, we obtain the following STL templates.

\noindent\textbf{Safety.} 
\noindent\textit{Torque limits:} For Barkour vb, the
actuators provide a peak output torque of 18Nm at each,
hence motor torque limits must be bounded by this limit. \\
\noindent\textit{Roll \& pitch limits:} To not loose the balance, roll and pitch of the robot's torso should be bounded. \\
\noindent\textit{CoM height:} Height of the torso should be lower-bounded to ensure no fall or tip over. \\
\noindent\textit{Support contacts:} We enhance the robot's stability by specifying the number of legs in stance according in a desired gait mode. Walk/trot requires $\ge2$ support contacts at all times (included only in the early curriculum stage and removed when training the bound gait.)  The shared safety specification for mode $g$ is
\[
\varphi_{\mathrm{safe},g}
=
\mathbf{G}_{\mathcal{W}}
\Big(
\rho_\tau\ge0 \wedge
\rho_{\phi,g}\ge0 \wedge
\rho_{\theta,g}\ge0 \wedge
\rho_{z,g}\ge0
\Big).
\]
\noindent\textbf{Velocity Tracking.}
The objective is for the robot to follow the commanded linear and angular velocities. 
The velocity tracking specification for mode g is
\[
\varphi_{\mathrm{track},g}
=
\mathbf{G}_{\mathcal{W}}
\Big(
\rho_{v_x,g}\ge0 \wedge
\rho_{v_y,g}\ge0 \wedge
\rho_{\omega,g}\ge0
\Big).
\]
\noindent\textbf{Walk-Trot.} 
This gait is characterized by support-rich diagonal locomotion with no flight. We use the following mode-specific predicates: \\
\noindent\textit{G1-Diagonal phase error:} This predicate is applied to both walk-trot and trot gaits, and it quantifies the temporal synchronization of diagonal leg pairs. \\
\noindent\textit{G2-Diagonal pair fraction:} This predicate quantifies how strongly the gait exhibits trot-like diagonal contact structure. \\
\noindent\textit{G3-Multi-support moment:} This predicate encourages the walk gait to exhibit at least one recent multi-support phase (three or more feet are simultaneously in contact) by rewarding the existence of a timestep within the trailing K-step window. The walk-trot specification is
\begin{equation}
\small
\varphi_W
=
\mathbf{G}_{\mathcal{W}}
\Big(
m_W\Rightarrow
(\rho_{\mathrm{diag},W}\ge0
\wedge \rho_{\mathrm{diag2},W}\ge0
\wedge \rho_{3+\mathrm{event}}\ge0)
\Big).
\end{equation}
\noindent\textbf{Trot Predicates:} 
While we have common predicates (G1 and G2) with walk-trot regime, trot is characterized by dominant diagonal 2-contact support and lower duty factor than walk-trot. \\
\noindent\textit{G4-Two-leg support:} This predicate is the fraction of timesteps in which exactly two feet are in contact with the ground. The trot specification is
\begin{equation}
\small
\varphi_T
=
\mathbf{G}_{\mathcal{W}}
\Big(
m_T\Rightarrow
(\rho_{\mathrm{diag},T}\ge0
\wedge \rho_{\mathrm{diag2},T}\ge0
\wedge \rho_{2\mathrm{contact}}\ge0)
\Big).
\end{equation}
\noindent\textbf{Bound Gait:} 
This is a high-speed pair-synchronized running gait where the forelegs and hind legs move approximately in phase, the two leg pairs alternate in stance, and locomotion includes one or more aerial phases, yielding larger sagittal extension and higher speed than diagonal trot. \\
\noindent\textit{B1-Minimum aerial-phase occupancy:} This computes the time-fraction when no foot is in contact with the ground, and enforces the minimum aerial-phase occupancy for bounding.\\
\noindent\textit{B2/3-Pair-only support states:} This computes the fractions of times when only the fore or hind legs are in contact; robustness enforces the minimum pair-only support occupancy for bounding.\\
\noindent\textit{B4-All four contact fraction:} This constrains the time fraction where all four legs are in contact. \\
\noindent\textit{B5-Trot suppression term:} This predicate computes the fraction of 2-contact states that correspond to diagonal leg pairs; thus, for the bound mode, positive robustness favors suppression of trot-like diagonal support patterns in favor of pair-based fore/hind support. \\
\noindent\textit{B6-Pair touchdown events:} This predicate enforce that there should be at least one front-pair touchdown event and at least one hind-pair touchdown event within the current temporal window. The bound specification is
\begin{equation}
\small
\begin{aligned}
\varphi_B
=
\mathbf{G}_{\mathcal{W}}
\Big(
m_B\Rightarrow
(\rho_{\mathrm{flight}}\ge0
\wedge \rho_{\mathrm{front}}\ge0
\wedge \rho_{\mathrm{hind}}\ge0 
&\wedge \\ \rho_{\mathrm{all4}}\ge0
\wedge \rho_{\mathrm{diag2},B}\ge0
\wedge \rho_{\mathrm{bound\_event}}\ge0)
&\Big).
\end{aligned}
\end{equation}

\noindent\textbf{Robustness Reward:} 
To obtain a smooth reward signal, we aggregate multiple robustness terms using a Boltzmann-weighted soft minimum 
$
\operatorname{smin}_{\beta}(z_1,\dots,z_K) $ 
$:= \sum_{k=1}^{K}\pi_k z_k$, where 
$
\pi_k = \frac{e^{-\beta z_k}}{\sum_{j=1}^{K} e^{-\beta z_j}}.
$
For the final reward shaping, each grouped robustness is normalized as $\tanh(\mu_i)$
to map to a value in $[-1,1]$, preventing large robustness magnitudes from dominating the reward.\footnote{All scale factors and weight terms are obtained by fitting the templates to expert datasets.} We also include the torque-effort penalty for energy efficiency
$
J_{\tau}(t)
=
\frac{1}{|W|}
\sum_{k\in W}
\sum_{j=1}^{12}
\tau_j(k)^2 .
$
The final scalar reward is
\begin{equation}
\begin{aligned}
r_t
=
&w_{\mathrm{safe}}^{(g)}\, \tanh(\mu_{\mathrm{safety}})
+
w_{\mathrm{track}}^{(g)}\, \tanh(\mu_{\mathrm{track}}) \\
+ 
&w_{\mathrm{pattern}}^{(g)}\, \tanh(\mu_{\mathrm{pattern}})
-
w_{\mathrm{effort}} J_{\tau}(t).
\end{aligned}
\end{equation}
where $w_i \in [1.0, 1.3]$ are corresponding weight terms.

\begin{table*}[t]
\centering
\small
\caption{Benchmark comparison over 20 rollouts per commanded velocity. CoT: lower is better; survival and success: higher is better. Green/red denote qualitative video-based assessment of mode-appropriate gait behavior. CoT sample standard deviations are \(\leq0.01\).}
\label{tab:benchmark_per_velocity}

\setlength{\tabcolsep}{2pt}
\renewcommand{\arraystretch}{1.08}

\begin{adjustbox}{max width=\textwidth}
\begin{tabular}{c@{\hspace{4pt}}c|ccc|ccc|ccc|ccc|ccc}
\toprule

& &
\multicolumn{3}{c|}{\textbf{Human-STL}}
& \multicolumn{3}{c|}{\textbf{GPT-STL}}
& \multicolumn{3}{c|}{\textbf{Expert Oracle}}
& \multicolumn{3}{c|}{\textbf{Heuristic}}
& \multicolumn{3}{c}{\textbf{Text2Reward}} \\

\(\mathbf{v_x}\)
& \textbf{Gait}

& CoT $\downarrow$
& Surv. $\uparrow$
& Succ. $\uparrow$

& CoT $\downarrow$
& Surv. $\uparrow$
& Succ. $\uparrow$

& CoT $\downarrow$
& Surv. $\uparrow$
& Succ. $\uparrow$

& CoT $\downarrow$
& Surv. $\uparrow$
& Succ. $\uparrow$

& CoT $\downarrow$
& Surv. $\uparrow$
& Succ. $\uparrow$ \\
\midrule

0.3 & Walk
& 2.1 & 100\% & \gaitok{100\%}
& 2.0 & 100\% & \gaitok{100\%}
& 0.9 & 100\% & \gaitok{100\%}
& 1.1 & 100\% & \gaitok{100\%}
& 1.1 & 100\% & \gaitok{100\%} \\

0.5 & Walk
& 1.5 & 100\% & \gaitok{100\%}
& 1.6 & 100\% & \gaitok{100\%}
& 0.9 & 100\% & \gaitok{100\%}
& 1.0 & 100\% & \gaitok{100\%}
& 0.8 & 100\% & \gaitok{100\%} \\

0.7 & Walk
& 1.3 & 100\% & \gaitok{100\%}
& 1.4 & 100\% & \gaitok{100\%}
& 1.0 & 100\% & \gaitok{100\%}
& 1.0 & 100\% & \gaitok{100\%}
& 0.9 & 100\% & \gaitok{100\%} \\

1.0 & Trot
& 1.2 & 100\% & \gaitok{100\%}
& 1.3 & 100\% & \gaitok{100\%}
& 1.1 & 100\% & \gaitok{100\%}
& 1.1 & 100\% & \gaitok{100\%}
& 1.1 & 100\% & \gaitok{100\%} \\

1.3 & Trot
& 1.1 & 100\% & \gaitok{100\%}
& 1.2 & 100\% & \gaitok{100\%}
& 1.3 & 100\% & \gaitok{100\%}
& 1.2 & 100\% & \gaitok{100\%}
& 1.1 & 100\% & \gaitbad{100\%} \\

1.6 & Trot
& 1.0 & 100\% & \gaitok{100\%}
& 1.3 & 100\% & \gaitbad{100\%}
& 1.3 & 100\% & \gaitok{100\%}
& 1.4 & 100\% & \gaitok{100\%}
& 1.0 & 100\% & \gaitbad{100\%} \\

1.9 & Bound
& 1.1 & 100\% & \gaitok{100\%}
& 1.3 & 100\% & 0\%
& 1.3 & 95\% & 0\%
& 1.4 & 100\% & \gaitbad{100\%}
& -   & 0\% & 0\% \\

2.0 & Bound
& 1.1 & 100\% & \gaitok{100\%}
& 1.3 & 100\% & 0\%
& 1.4 & 95\% & 0\%
& 1.4 & 100\% & 0\%
& - & 0\% & 0\% \\

2.1 & Bound
& 1.2 & 100\% & \gaitok{70\%}
& 1.3 & 100\% & 0\%
& 1.4 & 100\% & 0\%
& 1.4 & 100\% & 0\%
& - & 0\% & 0\% \\

\bottomrule
\end{tabular}
\end{adjustbox}

\vspace{-6pt}
\end{table*}

\noindent{\bf Data Collection \& Feature Extraction.}
To capture diverse locomotion behaviors, we compile trajectory datasets (50 expert trajectories per regime) from three specialized models corresponding to low-speed (walk-trot), mid-high speed (trot), high speed (bound) regimes. We adopt the base model from \cite{caluwaerts2023barkour}, introducing a revised reward formulation and a curriculum learning strategy for the different speed profiles. Although these parameters could be estimated using heuristics from the literature, we opt for this data-driven approach to directly leverage the available expert trajectories.

Each trajectory consists of \(500\) simulation steps with timestep \(\Delta t = 0.02\,\text{s}\). To remove startup transients, we discard an initial warmup interval \(t \in [0, t_{\text{warm}}]\), where \(t_{\text{warm}} = 100\) steps (\(\sim 2\,\text{s}\)). All features are computed from the post-warmup trajectory and include signals describing tracking performance, stability, contact coordination, and gait structure. \textbf{Tracking features:} $|v_x(t)-v_x^{cmd}(t)|$, $|v_y-\!v_y^{cmd}|$, and $|\omega_z-\omega_z^{cmd}|$, 
\textbf{safety/stability features:} Center of mass (CoM) position $p_{com}(t)$ and velocity $v_{com}(t)$, a center of pressure (CoP) proxy $p_{cop}(t)$, base roll/pitch $(\phi(t),\theta(t))$, CoP-CoM distance $\|p_{cop}(t)-p_{com}^{xy}(t)\|_2$, and slip proxy $\max_{\ell\in\mathcal{C}(t)}\|v_\ell^{xy}(t)\|_2$, \textbf{contact-pattern features:} Foot contacts $c_\ell(t)\in\{0,1\}$ for $\ell\in\{FL,FR,HL,HR\}$, pair-synchronization errors (diagonal pairs for trot, front/hind pairs for bound), \textbf{stride-level features:} estimated stride period $T$ and duty factor $DF$ from rising/falling edges of contact signals.
We define contact count $n_c(t)=\sum_\ell c_\ell(t)$ and foot tangential speed $\|v_\ell^{xy}(t)\|_2$ in stance.
For gait structure we compute windowed statistics such as stride period $\bar T(t)$, duty factor $\overline{DF}(t)$, diagonal phase error $e_{diag}(t)$, and the fraction of 2-contact states that correspond to diagonal pairs $p_{diag2}(t)$. 

We fit parameters of the PSTL templates from data in a quantile-based approach. For each atomic predicate $p_A(t)$ used in the template, we select thresholds based on empirical quantiles of the successful trajectories. For example, for a strict windowed-$\mathbf{G}$ implementation (min/softmin over time), we use tolerant high quantiles (e.g., 99th percentile for error magnitudes and a low percentile for floors) to reduce over-penalizing short transients.



\section{Experimental Results}

\noindent{\bf Robot and Simulation Setup.}
The locomotion controller is developed for the Barkour vb quadruped robot, with simulation performed in MuJoCo XLA (MJX), a JAX-based implementation of MuJoCo that enables highly parallelized GPU simulation \cite{caluwaerts2023barkour,todorov2012mujoco}.
We train the locomotion policy using Brax’s JAX-based PPO implementation as it is optimized for accelerator-parallel simulation \cite{ppo, kohl2004policy,freeman2021brax}. PPO is trained for $400$M timesteps using an unroll length of $30$, $32$ minibatches, $4$ updates per batch, a discount factor of $0.955$, a learning rate of $1.5\times10^{-4}$, an entropy cost of $0.004$, $8192$ parallel environments, and a batch size of $256$. All reward variants use the same policy architecture, PPO optimization settings, training budget, command distribution, and domain-randomization procedure; only the reward formulation differs across methods. Complete code and configurations will be released upon publication. During training, a new velocity command is randomly sampled every 10 s. The initial command range is set to $v_x^\star \in [0.0,\,1.7]\,{m/s}$,
$v_y^\star \in [-0.2,\,0.2]\,{m/s}$,
and $\dot\omega^\star \in [-0.2,\,0.2]\,{rad/s}$ for the walk/trot regimes. The forward-velocity range is then gradually expanded to $v_x^\star \in [0.0,\,1.9]\,{m/s}$ as part of the curriculum for learning the bound gait. In our work, we randomize two critical sets of physical parameters: friction (chosen in the range $\mathbf{(0.6, 1.4)}$) and actuator gain and bias (chosen in the range $\mathbf{(-5, 5)}$) as a part of domain randomization. This forces the learned policy to cope with slippery or sticky terrains and be robust to inaccurate torque control.

We compare the following methods: (1) \textbf{Human-STL} is our manually designed STL-based reward, (2) \textbf{GPT-STL} based on STL specifications generated by GPT 5.5 and serves as an ablation for whether the LLM can replace manual specification design \cite{openai2026gpt55}, (3) \textbf{Expert-switching oracle} selects the corresponding specialized expert for the ground-truth commanded speed regime, (4) \textbf{Heuristic} uses the standard hand-engineered locomotion reward in the original Barkour paper \cite{caluwaerts2023barkour}, 
(5) \textbf{Text2Reward} adapts direct LLM-generated reward code \cite{xie2024text2reward} (implemented using GPT-5.5 as the underlying language model).

GPT-STL follows the same pipeline as Human-STL for parameter grounding from expert trajectories, STL robustness computation, and PPO training. The key difference is the specification-design stage: instead of manually defining the PSTL templates, GPT-5.5 generates their symbolic structure from the locomotion task description and constrained STL grammar. 

We evaluate each policy under forward velocity commands
$
v_x$ $\in$ $\{0.3,0.5,0.7,1.0,1.3,$ $1.6,1.9,2.0,2.1\}$ m/s,
with zero lateral and yaw commands. For each command, we run
$20$ independent rollouts, each with a horizon of $500$ steps.
For command-tracking performance, we ignore the first $50$ steps as
a warm-up period and evaluate tracking only over the remaining trajectory. We leverage three evaluation metrics: {\bf cost of transportation (CoT)}, {\bf survival rate}, and {\bf success rate}. We measure energy efficiency using cost of transportation $\frac{E}{m\,g\,d}$ where $E$ is the total energy consumed, $m$ is the robot mass,
$g$ is gravitational acceleration, and $d$ is the planar distance traveled during the rollout. 

Lower CoT indicates better transport efficiency. A rollout is counted as \emph{survived} if the robot does not terminate early
due to falling or violating the environment termination conditions.
We define success based on velocity tracking accuracy after the warm-up phase.
Let $v_{x,\mathrm{cmd}}$ be the commanded forward velocity, and let
$v_{x,\mathrm{loc}}(t)$ denote the robot's forward velocity in its local body
frame. The average post-warm-up forward speed for rollout $k$ is
$
\bar{v}_x^{(k)}
=
\frac{1}{T-50}
\sum_{t=51}^{T}
v_{x,\mathrm{loc}}^{(k)}(t).
$
A rollout is considered successful if
$
\left|\bar{v}_x^{(k)} - v_{x,\mathrm{cmd}}\right|
\leq 0.15\,|v_{x,\mathrm{cmd}}|,
$
i.e., the average forward speed stays within $\pm 15\%$ of the commanded
velocity. 

Gait behavior is additionally assessed qualitatively through inspection of rollout videos. We determine whether the observed leg-pair coordination and support/aerial-phase structure are consistent with the intended walk, trot, or bound regime. This qualitative annotation is used only to characterize gait behavior and is separate from the quantitative success metric defined below. Representative rollouts for all methods and gait regimes are provided in the accompanying video.

\begin{table}[t]
\centering
\caption{
Ablation over temporal window size $H$. Metrics are averaged across all evaluated commanded forward velocities. Green indicates that the expected gait is achieved, while red indicates that it is not.
}
\label{tab:window_ablation_mg}
\setlength{\tabcolsep}{7.0pt}
\renewcommand{\arraystretch}{1.05}
\begin{tabular}{@{}|l | c | c c c |@{}}
\toprule
Method & $H$ & CoT $\downarrow$ & Survival $\uparrow$
& Success $\uparrow$ \\
\midrule

\multirow{5}{*}{GPT-STL}
& 1  & 1.4 & 100\% & \textcolor{BrickRed}{55.6\%}  \\
& 5  & 1.6 & 100\% & \textcolor{BrickRed}{55.6\%}  \\
& 10 & 1.4 & 100\% & \textcolor{BrickRed}{55.6\%}  \\
& 20 & 1.4 & 100\% & \textcolor{BrickRed}{66.7\%}  \\
& 30 & 1.4 & 99.4\% & \textcolor{BrickRed}{66.7\%}  \\

\midrule

\multirow{5}{*}{Human-STL}
& 1  & $1.3$ & $100\%$ & \textcolor{BrickRed}{$100\%$} \\
& 5  & $1.1$ & $99.4\%$ & \textcolor{BrickRed}{$99.4\%$}  \\
& 10 & $1.4$ & $100\%$ & \textcolor{ForestGreen}{$100\%$} \\
& 20 & $1.3$ & $99.4\%$ & \textcolor{ForestGreen}{$99.4\%$} \\
& 30 & $1.3$ & $100\%$ & \textcolor{ForestGreen}{$95.6\%$} \\

\bottomrule
\end{tabular}
\vspace{-10pt}
\end{table}

\section{Discussion}
\label{sec:discussion}

The benchmark results reveal an important distinction between task-level
success and locomotion behavior. Human-STL is the only method that consistently
realizes the expected gait across the evaluated velocity range, while
maintaining $100\%$ success up to $2.0\,\mathrm{m/s}$ and $70\%$ success at
$2.1\,\mathrm{m/s}$. In contrast, both Text2Reward and the heuristic baseline
can satisfy the velocity-based success criterion without producing the desired
gait. Text2Reward, for example, achieves $100\%$ success at $1.3$ and
$1.6\,\mathrm{m/s}$ despite exhibiting an incorrect gait, while the heuristic
policy achieves $100\%$ success at $1.9\,\mathrm{m/s}$ without transitioning
to bound. This exposes a limitation of evaluating locomotion primarily through
command tracking: matching forward velocity does not imply that the policy has
learned the intended temporal organization of contacts. CoT should similarly
be interpreted cautiously at very low velocities, where the small traveled
distance can inflate the normalized value. More importantly, these results
show that STL adds information that is not captured by either velocity success
or energetic efficiency.

\begin{table}[t]
\centering
\caption{Transition success rates under different temporal windows $H$.
Each transition changes the commanded forward velocity from $v_1$ to $v_2$. Green indicates that the expected gait is achieved, while red indicates that it is not.}
\label{tab:transition_results}
\small
\setlength{\tabcolsep}{3pt}
\renewcommand{\arraystretch}{1.0}

\begin{tabular}{lccccc}
\toprule
\textbf{Transition} &
\textbf{$v_1$ (m/s)} &
\textbf{$v_2$ (m/s)} &
\textbf{$H=10$} &
\textbf{$H=20$} &
\textbf{$H=30$} \\
\midrule

Walk $\rightarrow$ Trot
    & 0.3 & 0.8 & 100\% & 100\% & 95\% \\
Walk $\rightarrow$ Trot
    & 0.3 & 1.2 & 100\% & 100\% & 100\% \\
Walk $\rightarrow$ Trot
    & 0.6 & 0.8 & 100\% & 100\% & 100\% \\
Walk $\rightarrow$ Trot
    & 0.6 & 1.2 & 100\% & 100\% & 100\% \\

\midrule
Trot $\rightarrow$ Walk
    & 0.8 & 0.3 & 100\% & 100\% & 100\% \\
Trot $\rightarrow$ Walk
    & 1.2 & 0.3 & 100\% & 100\% & 100\% \\
Trot $\rightarrow$ Walk
    & 0.8 & 0.6 & 100\% & 100\% & 100\% \\
Trot $\rightarrow$ Walk
    & 1.2 & 0.6 & 100\% & 100\% & 100\% \\

\midrule
Trot $\rightarrow$ Bound
    & 1.0 & 1.75 & \textcolor{ForestGreen}{90\%} & \textcolor{BrickRed}{15\%} & \textcolor{BrickRed}{25\%} \\
Trot $\rightarrow$ Bound
    & 1.0 & 1.9  & \textcolor{ForestGreen}{70\%} & \textcolor{BrickRed}{30\%} & \textcolor{ForestGreen}{35\%} \\
Trot $\rightarrow$ Bound
    & 1.5 & 1.75 & \textcolor{BrickRed}{50\%} & \textcolor{BrickRed}{15\%} & 0\% \\
Trot $\rightarrow$ Bound
    & 1.5 & 1.9  & \textcolor{ForestGreen}{55\%} & 0\%  & \textcolor{BrickRed}{5\%} \\

\midrule
Bound $\rightarrow$ Trot
    & 1.75 & 1.0 & \textcolor{BrickRed}{10\%} & \textcolor{ForestGreen}{100\%} & \textcolor{ForestGreen}{50\%} \\
Bound $\rightarrow$ Trot
    & 1.9  & 1.0 & \textcolor{BrickRed}{10\%} & \textcolor{ForestGreen}{85\%}  & \textcolor{BrickRed}{55\%} \\
Bound $\rightarrow$ Trot
    & 1.75 & 1.5 & 0\%  & \textcolor{ForestGreen}{85\%}  & \textcolor{ForestGreen}{20\%} \\
Bound $\rightarrow$ Trot
    & 1.9  & 1.5 & 0\%  & \textcolor{ForestGreen}{85\%}  & \textcolor{BrickRed}{40\%} \\

\bottomrule
\end{tabular}
\vspace{-15pt}
\end{table}

The comparison between Human-STL and GPT-STL helps isolate where the remaining
difficulty lies. GPT-STL achieves reliable walking and trotting, but its gait
behavior starts to deviate at the upper end of the trot regime. Human-STL, using the same
learning framework but manually specified temporal structure, successfully
extends to the bound regime. In this sense, STL provides a useful intermediate
representation between language and control: unlike direct reward generation,
the generated object has an interpretable temporal meaning, and failures can be
explained to missing or inaccurate behavioral constraints. The gap between
GPT-STL and Human-STL therefore also provides a concrete target for improving
future LLM-based specification generation.

The Expert Oracle result provides additional context for the high-speed regime.
Bound expert trajectories were collected only over
$[1.5,1.8]\,\mathrm{m/s}$, since reliable expert trajectories could not be
obtained at higher commanded velocities. Its failure at
$1.9$--$2.1\,\mathrm{m/s}$ should therefore be interpreted as extrapolation
beyond the demonstration range rather than as evidence that no bound expert
behavior was available. Interestingly, the unified Human-STL policy remains
successful beyond this range. This indicates that the learned temporal reward
is not restricted to reproducing the exact velocities observed in the expert
data; once the relevant gait structure is encoded, reinforcement learning can
extend that structure to commands outside the expert demonstration range.

The horizon ablation further supports the role of temporal structure. For
Human-STL, aggregate success changes modestly across window sizes:
$H=1$ achieves $100\%$ success, \\ while $H=30$ still achieves $95.6\%$.
However, the behavioral outcome changes qualitatively. The short windows
$H=1$ and $H=5$ do not consistently recover the expected gait, whereas
$H\geq10$ does. Thus, the value of a temporal horizon is not primarily an
improvement in conventional task success; it allows the reward to distinguish
between trajectories that have similar instantaneous velocity but different
contact evolution over time. This is precisely the setting in which temporal
logic is useful for locomotion.

The transition experiments reinforce this interpretation while also showing
that different gait changes require different temporal context. Walk--trot and
trot--walk transitions are nearly perfect across all tested horizons, whereas
transitions involving bound are substantially more sensitive. Trot-to-bound
transitions are generally strongest with $H=10$, while bound-to-trot
transitions are most reliable with $H=20$. Rather than implying that a longer
or shorter horizon is uniformly preferable, this asymmetry suggests that the
relevant temporal context depends on the transition being executed. Entering a
highly dynamic gait and settling back into a slower gait impose different
temporal requirements. Together with the horizon ablation, these results
suggest that an adaptive or transition-conditioned STL horizon could be a
natural extension of the current formulation.

\section{Conclusion}
\label{sec:conclusion}

We introduced an STL-based framework for learning a unified quadruped
locomotion policy across multiple gait regimes and gait transitions. The
results show that explicitly representing temporal locomotion structure is
useful beyond velocity tracking alone: STL formulation provides
a direct mechanism for encoding and evaluating the desired gait behavior.

Our study is conducted entirely in simulation using MJX. We apply domain
randomization over friction and actuator parameters and additionally evaluate
the learned policy in terrain settings that were not the primary target of the
training setup. Despite not explicitly optimizing for uneven-terrain
locomotion, the policy succeeds in $17/20$ trials on a $5^\circ$ downhill
slope, $9/20$ on a $5^\circ$ uphill slope, $11/20$ on shallow descending
stairs, and $4/20$ on shallow ascending stairs. These results are encouraging,
particularly given that terrain geometry is not currently represented in the
gait-selection mechanism. Incorporating terrain information such as
slope and roughness into mode selection and specification generation is a
natural next step, together with validation beyond simulation.

\begin{figure}[t]
    \centering
    \includegraphics[width=\linewidth]{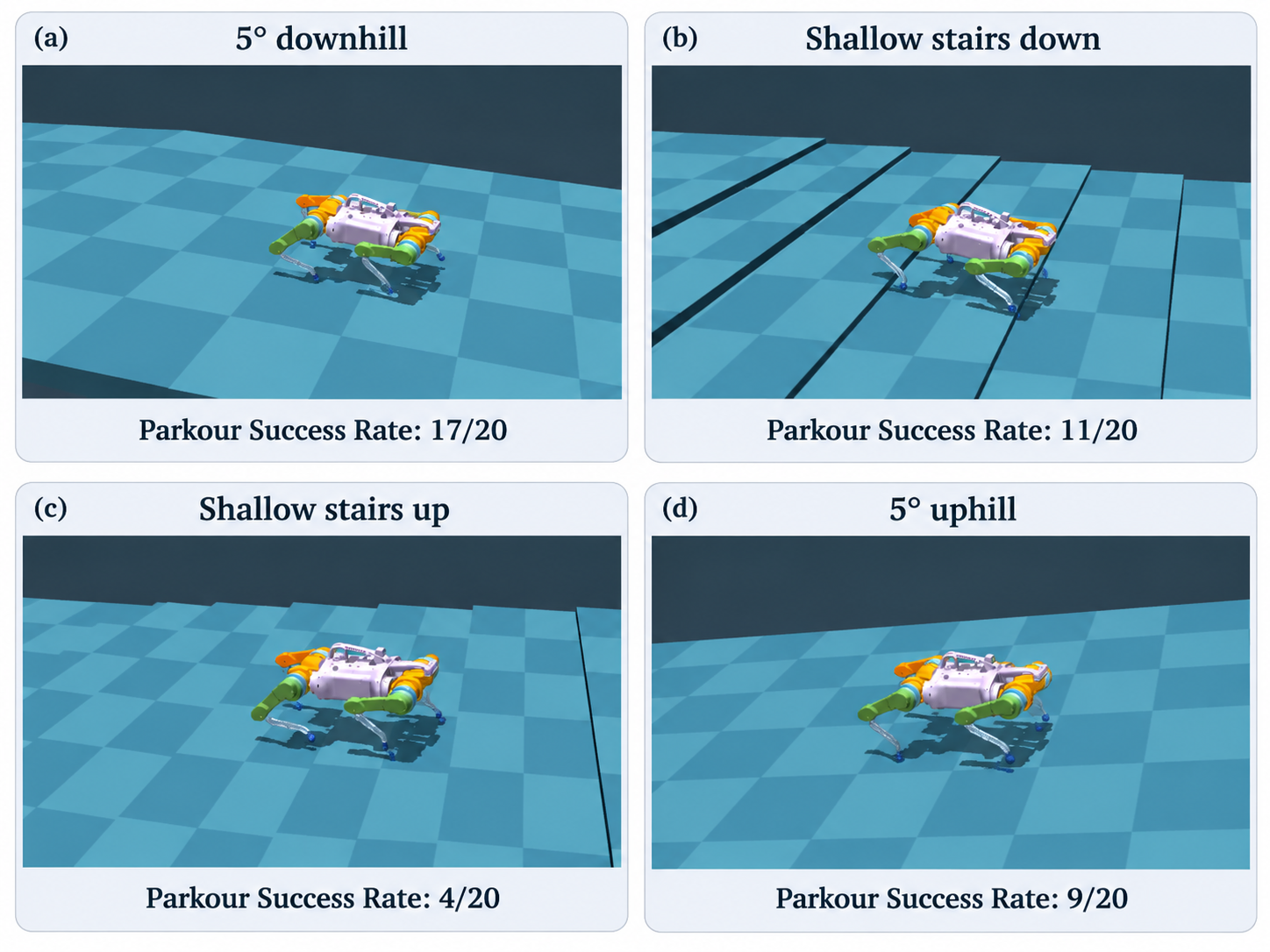}
    \caption{Uneven Terrain Results.}
    \label{fig:terrains}
     \vspace{-15pt} 
\end{figure}


\bibliographystyle{IEEEtran}
\bibliography{ref,IEEEabrv}

\end{document}